\documentclass[conference]{IEEEtran}
\IEEEoverridecommandlockouts
\usepackage{comment}
\usepackage{listings}
\usepackage{microtype}
\usepackage{graphicx}
\usepackage{balance}
\usepackage[T1]{fontenc}
\usepackage{enumitem}
\usepackage{colortbl}
\usepackage{dblfloatfix}
\usepackage{booktabs,tabularx,multirow}
\usepackage{cite}
\usepackage{amsmath,amssymb,amsfonts}
\usepackage{algorithm}
\usepackage{algpseudocode}  
\usepackage{textcomp}
\usepackage{xcolor}
\usepackage[most]{tcolorbox}
\def\BibTeX{{\rm B\kern-.05em{\sc i\kern-.025em b}\kern-.08em
    T\kern-.1667em\lower.7ex\hbox{E}\kern-.125emX}}

\title{Structured Authority Injection: Persistent Human Control for LLM Agents via Typed MCP Overrides}

\author{
\IEEEauthorblockN{
Ryan Tran, 
YueMing Zhang
}
\IEEEauthorblockA{
California State University Long Beach\\
Long Beach, CA, USA\\
\{ryan.tran05, simon.zhang01\}@student.csulb.edu
}

\and
\IEEEauthorblockN{Lingfeng Tao}
\IEEEauthorblockA{
Kennesaw State University\\
Kennesaw, GA, USA\\
ltao2@kennesaw.edu
}

\and

\IEEEauthorblockN{Nan Yang}
\IEEEauthorblockA{
University of Sydney\\
Sydney, NSW, Australia\\
n.yang@sydney.edu.au
}

\and

\IEEEauthorblockN{Shengjie Xu}
\IEEEauthorblockA{
University of Arizona\\
Tucson, AZ, USA\\
sjxu@arizona.edu
}

\and

\IEEEauthorblockN{
Yu Yin
}
\IEEEauthorblockA{
Case Western Reserve University\\
Cleveland, OH, USA\\
yu.yin@case.edu
}
\and

\IEEEauthorblockN{
Shuai Xu
}
\IEEEauthorblockA{
Case Western Reserve University\\
Cleveland, OH, USA\\
shuai.xu2@case.edu
}

\and
\IEEEauthorblockN{
Wenlu Zhang,
Hailu Xu
}
\IEEEauthorblockA{
California State University Long Beach\\
Long Beach, CA, USA\\
\{wenlu.zhang, hailu.xu\}@csulb.edu
}

}

\begin{document}

\maketitle
\begingroup

\renewcommand\thefootnote{}
\footnotetext{Corresponding E-mail: hailu.xu@csulb.edu}
\addtocounter{footnote}{-1}
\endgroup

\begin{abstract}
LLM-based agents are increasingly used as autonomous orchestrators in distributed and multi-agent systems, yet a control gap remains: no protocol-level mechanism allows human operators to inject binding directives that persist across planning cycles. Free-text overrides suffer from ambiguous interpretation due to lack of schema validation and from authority erosion as operator instructions lose influence within accumulating context. We propose \textbf{Structured Authority Injection (SAI)}, a protocol that addresses these issues by routing operator corrections through typed, schema-validated Model Context Protocol (MCP) tool calls and maintaining a ranked authority context prepended to each LLM prompt. SAI defines three override classes (Constraint, Priority, and Strategy) with distinct persistence semantics and an explicit precedence order enforced via a directed authority channel inaccessible to the planning process. The method requires no model fine-tuning or access to model weights and applies to instruction-tuned LLMs supporting MCP tool use. Evaluation across nine LLMs and eight scenarios targeting agent failure modes shows reduced ambiguous directive interpretation and mitigated authority decay across planning iterations, with lower prompt overhead in the strongest models (up to 38\% depending on model and scenario), though benefit varies substantially across models. SAI provides a general design pattern for agent systems requiring explicit and persistent human control.

\end{abstract}


\section{Introduction}
\label{sec:intro}
As large language models (LLMs) mature from conversational tools into autonomous
planning agents, they are increasingly positioned as high-level orchestrators in
distributed systems, such as allocating computational resources, coordinating
specialized sub-agents, and executing adaptive action sequences across
heterogeneous infrastructure~\cite{kannan2024smartllm, liang2023code, zhang2026swiftbot, ahn2022saycan}.
This shift from passive assistant to active agent introduces a governance
challenge that existing system designs have not resolved: \textit{how can a
human operator inject binding, persistent directives into a live LLM-driven
orchestration loop without modifying model weights, restarting the planning
pipeline, or relying on ad~hoc prompt conventions?}
 
The Model Context Protocol (MCP)~\cite{anthropic_mcp_2024} has emerged as a standardized
interface that enables LLMs to invoke external tools through schema-validated,
auditable function calls, extending the reach of agentic systems to real-world
services and infrastructure~\cite{hou2025model, fu2025rosbag, guo2025lightllm}.
In current deployments, however, MCP functions exclusively as a \textit{tool
bus}: the LLM selects and calls tools, and results flow back into the planning
context.
Human operators who need to correct an in-flight decision---redirecting a
misallocated agent, elevating the urgency of a time-critical job, or imposing a
constraint absent from the world-state snapshot---have no structured channel for
doing so.
They are reduced to appending free-text messages to the conversation, a practice
that introduces two structural deficiencies.

\smallskip
\noindent\textbf{Deficiency~1: Ambiguous uptake.}
Natural-language corrections carry no schema and admit no validation~\cite{shi2023distracted}.
The LLM must infer the target entity, constraint scope, and temporal validity
from unstructured text.
Instruction-tuned models may partially acknowledge such feedback while reverting
to prior planning priors~\cite{huang2022inner,shi2023distracted}, producing
silently non-compliant behavior that is difficult to detect without exhaustive
output inspection.
 
\smallskip
\noindent\textbf{Deficiency~2: Authority erosion.}
Even correctly interpreted corrections lose influence over
time~\cite{liu2024lost}.
As new observations and tool-call results accumulate in the context window,
earlier operator directives compete on equal footing with non-binding
conversational content.
Positional bias in transformer attention~\cite{liu2024lost,wang2023primacy}
causes late-context instructions to carry systematically lower weight, eroding
the authority of operator directives across successive planning epochs without
any explicit override or cancellation.
 
\smallskip
Both deficiencies stem from a shared architectural limitation: the LLM context
window is a flat, undifferentiated stream that assigns no structural privilege
to human-issued directives over model-generated content.

In this paper we present \textbf{Structured Authority Injection (SAI)}, a
protocol that transforms MCP from a bidirectional tool bus into a
\textit{one-directional authority channel}.
SAI routes operator corrections through typed MCP tool calls whose schemas are
enforced server-side before any record reaches the planning agent, and maintains
a persistent, ranked authority context prepended to every subsequent LLM prompt.
The protocol makes three contributions:

\begin{enumerate}
 
\item \textbf{Override taxonomy.}
Three semantically distinct correction classes (Constraint, Priority,
Strategy), each with typed schemas, class-specific persistence horizons, and
deterministic conflict resolution rules (Section~\ref{sec:taxonomy}).
 
\item \textbf{MCP as a human authority layer.}
MCP's schema-validation machinery is inverted to give human supervisors
tamper-resistant write-access to the LLM's planning context.
Override tools are callable exclusively by the supervision interface, never by
the planning agent, making the authority channel architecturally resistant to
adversarial prompting or model misbehavior.
(Section~\ref{sec:schemas}).
 
\item \textbf{Systematic empirical evaluation.}
Nine LLMs spanning proprietary and open-weights families are evaluated across eight scenarios, each isolating a distinct agent failure
mode.
SAI repairs task completion for partially failing models and reduces token consumption by up to 38\% in the strongest models, while revealing that its benefit is highly model-dependent and not predicted by model scale.
 
\end{enumerate}
 
While the evaluation instantiates SAI in a multi-agent task-allocation setting,
the protocol is domain-agnostic: its authority channel design applies equally to
LLM-driven network orchestration, autonomous cloud resource management, and
AI-assisted security response, as any system context in which human operators must
enforce binding directives over a live agentic planning loop.

The following paper is organized as follows: In Section~\ref{sec:related} we introduce the related work. In Section~\ref{sec:sai}, we present the design and workflow of SAI. We perform the evaluation of eight scenarios in Section~\ref{sec:evaluation}. We discuss the design trade-offs and future work in Section~\ref{sec:discussion}, and summarize the conclusion in Section~\ref{sec:conclusion}.

\section{Related Work}
\label{sec:related}

\subsection{LLM-Driven Multi-Robot Collaboration}

Recent work has shown that LLMs can serve as high-level planners for multi-robot task allocation~\cite{zhang2025survey}.
Systems such as COHERENT~\cite{liu2024coherent} decompose missions into sub-tasks and assign them via iterative planning loops, while RoCo~\cite{mandi2024roco} and consensus-based methods~\cite{chen2024scalable} leverage multi-agent dialogue for assignment agreement.
MetaGPT~\cite{hong2024metagpt} further structures collaboration via role-based workflows and standardized operating procedures.
Other approaches, including SMART-LLM~\cite{kannan2024smartllm} and Code-as-Policies~\cite{liang2023code}, explore LLMs for coalition formation and executable policy generation.
Grounding methods such as SayCan~\cite{ahn2022saycan} and SayPlan~\cite{rana2023sayplan} additionally constrain LLM outputs using physical affordances or scene representations.

\subsection{Human-in-the-Loop Robot Systems}

Human-in-the-loop control has long been studied in shared autonomy, including mixed-initiative~\cite{heger2006sliding} and sliding autonomy frameworks~\cite{sellner2006sliding}, where control is dynamically shared between humans and robots.
Language-based correction mechanisms such as LILAC enable natural-language feedback for manipulation, though without explicit protocol structure~\cite{cui2023language}.
More recent LLM-based systems incorporate human feedback into multi-robot planning loops. HMCF~\cite{li2025hmcf} and Sirius~\cite{liu2023sirius} integrate operator corrections into plan revision or policy updates, while BrainBody-LLM~\cite{bhat2025brainbody} uses closed-loop human signals to trigger replanning in embodied tasks.
However, across these systems, human feedback is typically represented as unstructured language or control signals, lacking semantic parsing or protocol-level validation, which limits persistence and compositionality across planning steps.

\subsection{MCP in Robotic Systems}

The Model Context Protocol (MCP)~\cite{anthropic_mcp_2024} provides a standardized interface between LLMs and external tools.
In robotics, ROS-based MCP wrappers~\cite{fu2025rosbag, zimmer2024rosllm} expose robot control and sensing interfaces as structured tools, while IoT-MCP~\cite{yang2025iotmcp} extends this paradigm to heterogeneous sensor networks.
MCP-Zero~\cite{fei2025mcpzero} further enables dynamic tool discovery with reduced token overhead, and recent work has also highlighted security and access-control challenges in MCP-enabled agentic systems~\cite{hou2025model}.

\section{Structured Authority Injection}
\label{sec:sai}
 
\begin{figure*}[!t]
\centering
\includegraphics[scale=0.68]{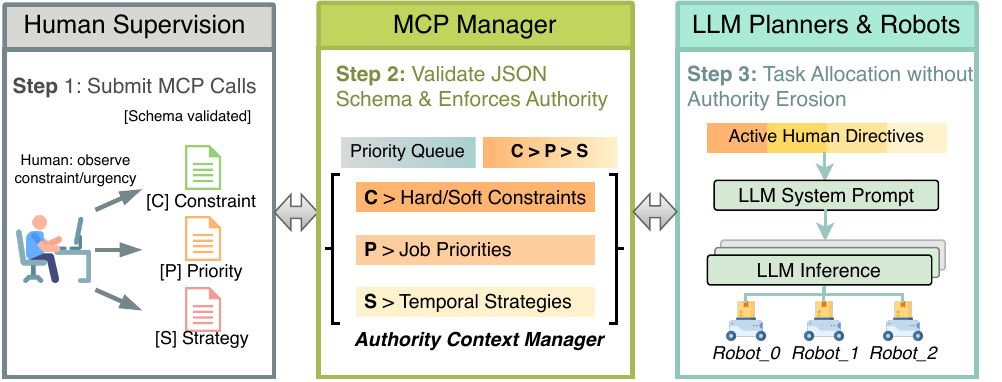}
\caption{The workflow and architecture of Structured Authority Injection (SAI): human supervisors submit schema-validated MCP override calls (Constraint, Priority, and Strategy), which are verified and persisted by the authority-context manager under deterministic precedence rules ($C \succ P \succ S$), then prepended to every subsequent LLM planning prompt to prevent authority erosion during multi-robot task allocation.}
\label{fig:mcp}
\end{figure*}

SAI is a \emph{protocol} that requires no fine-tuning, no access to model weights,
and no changes to the LLM's tool vocabulary.
It operates entirely at the interface between the human
supervision layer and the LLM prompt (as shown in Figure~\ref{fig:mcp}), exploiting two
properties of instruction-tuned models that are already
present: (i) the ability to follow typed tool schemas, and
(ii) the tendency to attend more strongly to early context~\cite{liu2024lost, wang2023primacy}.
The three components of SAI---the override taxonomy, the
MCP tool schemas, and the persistence model---are
interdependent and together constitute the protocol.
 
\subsection{SAI Taxonomy}
\label{sec:taxonomy}

A central design decision in SAI is to treat operator corrections not as an undifferentiated message stream, but as typed objects whose structure encodes the \emph{kind} of authority being exercised.
We identify three dimensions along which corrections differ fundamentally: (a)~the \emph{epistemic source}: what the operator knows that the LLM does not; (b)~the \emph{affected scope}: a single robot, a task class, or the whole fleet; and
(c)~the \emph{persistence horizon}: how long the correction should constrain future planning.
Conflating these dimensions forces the LLM to infer all three from free text, introducing the ambiguity failure mode~\cite{shi2023distracted}.

Analysis of failure events logged across our eight evaluation scenarios, combined with a review of the HRI supervisory-control literature~\cite{goodrich2008human, sellner2006sliding}, yields three semantically distinct and mutually exclusive correction classes that together cover the full observed correction space.

\noindent\textbf{Class~C (Constraint).} 
Encodes physical or operational limits absent from the world-state snapshot.
The planning agent's world model derives entirely from the JSON state serialized at each epoch; it has no access to out-of-band signals such as radio alerts, sensor failures, or direct operator observation.
Constraint overrides inject this knowledge as hard or soft rules scoped to a robot, a task, or a spatial region:
\begin{tcolorbox}[
    colback=gray!8,
    boxrule=0pt,
    frame hidden,
    left=4pt,
    right=4pt,
    top=4pt,
    bottom=4pt
]
\ttfamily\small
[HARD] Robot-2 battery is critical; do not assign further jobs.

[SOFT] Avoid the north corridor if possible.
\end{tcolorbox}
Hard constraints must never be violated; soft constraints are
bypassed only when no compliant assignment exists.
Both subtypes persist until the operator explicitly lifts them.

\noindent\textbf{Class~P (Priority).}
Encodes urgency information received through channels outside the
simulation, such as field radio calls, wearable sensor readings, medical
status updates~\cite{li2025hmcf}.
A priority override promotes a single task or an entire task class
to a new urgency level:
\begin{tcolorbox}[
    colback=gray!8,
    boxrule=0pt,
    frame hidden,
    left=4pt,
    right=4pt,
    top=4pt,
    bottom=4pt
]
\ttfamily\small
Job-6 is a confirmed cardiac arrest; treat as priority~5.
\end{tcolorbox}
The \texttt{scope} parameter (e.g., \texttt{this\_job} or
\texttt{job\_type}) determines whether the adjustment applies to
one instance or to all tasks of that type.
Priority overrides expire automatically when the affected task or
class is marked complete, requiring no manual cancellation.

\noindent\textbf{Class~S (Strategy).}
Encodes coordination heuristics that should govern robot
behavior over a bounded horizon, including zone distribution,
formation sequencing, concurrency limits~\cite{chen2024scalable}:

\begin{tcolorbox}[
    colback=gray!8,
    boxrule=0pt,
    frame hidden,
    left=4pt,
    right=4pt,
    top=4pt,
    bottom=4pt
]
\ttfamily\small
Keep robots in separate quadrants to maximize coverage.\\
Sequence the cargo formations; do not attempt both simultaneously.
\end{tcolorbox}

Strategy overrides are the most abstract class and carry a
\texttt{scope} parameter (e.g., \texttt{mission} or
\texttt{next\_N}) that triggers automatic expiry
after exactly $N$ planning cycles.
This temporal scoping is a deliberate usability choice:
operators under high cognitive load should not be required to
remember to cancel time-limited directives~\cite{goodrich2008human}.

The three classes impose a strict conflict-resolution order $\text{C} \succ \text{P} \succ \text{S}$. Within a class, a later override supersedes the earlier one for the same robot or task identifier without canceling overrides on different identifiers.
This precedence rule is evaluated deterministically at prompt construction time, adding no LLM inference to the planning loop.

\subsection{MCP Tool Schemas and Persistence Model}
\label{sec:schemas}

The taxonomy in Section~\ref{sec:taxonomy} defines \emph{what}
each correction class means; the MCP tool schemas and the
persistence model define \emph{how} corrections enter and persist
in the agent's planning context.
These two mechanisms jointly eliminate the ambiguity and
authority-erosion failure modes.

\begin{algorithm}[t]
\caption{SAI authority-context lifecycle (per epoch).}
\label{alg:sai}
\small
\begin{algorithmic}[1]
\Require Record set $\mathcal{A}$;
         override $o$ (if any);
         world-state $W$; history $H$
\Ensure Updated $\mathcal{A}$; tool calls $C$

\Statex \textit{\% Phase 1: override ingestion}
\If{$o$ received from supervision interface}
  \State \textbf{assert} valid caller credential
         \textbf{and} JSON Schema
         \Comment{else return error; LLM never sees $o$}
  \State $\mathcal{A} \leftarrow \mathcal{A} \cup \{o\}$
  \State Supersede earlier record in $\mathcal{A}$ with same
         class \textbf{and} entity identifier as $o$
\EndIf

\Statex \textit{\% Phase 2: prompt construction and planning}
\State Sort $\mathcal{A}$: $\text{C} \succ \text{P} \succ \text{S}$
\State $P \leftarrow [\textsc{SysPrompt}]
       \oplus \textsc{Render}(\mathcal{A})
       \oplus [W] \oplus [H]$
\State $C \leftarrow \textsc{LLM}(P)$

\Statex \textit{\% Phase 3: record expiry}
\ForAll{$a \in \mathcal{A}$}
  \If{$a.\textit{class} \in \{\texttt{C}, \texttt{P}\}$
      \textbf{and} \textsc{JobComplete}($a$)}
    \State $\mathcal{A} \leftarrow \mathcal{A} \setminus \{a\}$
           \Comment{job resolved or operator-cancelled}
  \ElsIf{$a.\textit{class} = \texttt{S}$
         \textbf{and} $a.\textit{scope} = \texttt{next\_N}$}
    \State $a.r \leftarrow a.r - 1$
    \If{$a.r = 0$}
      \State $\mathcal{A} \leftarrow \mathcal{A} \setminus \{a\}$
             \Comment{auto-expire after $N$ cycles}
    \EndIf
  \Else
    \State \textbf{skip}
           \Comment{$\texttt{mission}$-scoped S: persists until end}
  \EndIf
\EndFor

\State \Return $\mathcal{A},\, C$
\end{algorithmic}
\end{algorithm}

\textit{\textbf{Schema-validated tool calls.}}
Each override class is exposed as a typed MCP tool~\cite{anthropic_mcp_2024}.
The schemas enforce three design principles:
(i)~every parameter that affects persistence, scope, or precedence
is explicit and typed---never inferred from free text;
(ii)~enumerations constrain all bounded value spaces
(\texttt{severity}, \texttt{scope}, \texttt{new\_priority});
(iii)~free-text fields are restricted to human-readable rationale
slots and carry no structural parameters.

Schema validation runs server-side before any record reaches the
planner~\cite{hou2025model}.
Invalid submissions, such as a missing \texttt{robot\_id}, an
out-of-range priority, or an ill-formed scope, are rejected
immediately with a structured error returned to the supervision
interface, prompting the operator to correct the input.
The LLM therefore never receives a malformed override
record: the validation firewall is architecturally prior to the
prompt pipeline, eliminating the ambiguity failure mode at
submission time rather than at inference time.

\textit{\textbf{Asymmetric callability.}}
A role-checking middleware in the MCP server enforces that SAI
override tools are callable only by the supervision
interface, identified by a trusted caller credential attached to
each request.
The planning agent can observe active override records in its
authority context but cannot create, modify, or delete them.

\textit{\textbf{Persistent authority context.}}
To eliminate authority erosion, SAI maintains a
persistent authority context (an ordered list of active
override records), that is prepended to every subsequent
LLM prompt.
Each record stores the override class, validated parameters, a
one-sentence human-readable summary, a creation timestamp, and
(for \texttt{next\_N} overrides) a
remaining-count $r \in \mathbb{N}$ decremented after each
planning cycle.

The context is rendered as a numbered directive list and inserted at the very beginning of the system prompt, before the world-state snapshot and tool-call history. This placement exploits the primacy effect in instruction-tuned
LLMs~\cite{wang2023primacy, liu2024lost}: early-context tokens receive
disproportionately high attention weights.
Together, positional primacy and explicit precedence instruction
ensure that human directives structurally outrank model priors
without fine-tuning or retrieval-augmented memory.

Records expire by one of three routes: constraint and priority
overrides persist until the operator cancels them or
job completion fires for the affected entity;
mission-scoped strategy overrides persist until end of mission;
next\_N-scoped overrides are removed
automatically when their remaining-count reaches zero.
Algorithm~\ref{alg:sai} formalizes the complete per-epoch
lifecycle (override ingestion, prompt construction, agent
planning, and record expiration) as a single coherent procedure.

\subsection{Theoretical Properties}
\label{sec:properties}
 
We state three properties of SAI that distinguish it
from free-text override approaches and motivate its use
as a general protocol.
 
\noindent
\textbf{Property 1 (Submission-time validity).}
Every override record that enters the authority context
has been validated against its class schema.
The LLM never encounters a structurally incomplete or
type-invalid directive.
 
\noindent
\textbf{Property 2 (Tamper resistance).}
The LLM cannot modify the authority context.
Because SAI override tools are callable only by the
supervision interface, a misbehaving or adversarially
prompted model cannot inject, escalate, or remove
directives~\cite{hou2025model}.
This property holds as long as the MCP server correctly
enforces caller credentials.
 
\noindent
\textbf{Property 3 (Deterministic conflict resolution).}
Conflicts between override records are resolved by
static precedence, not by LLM inference.
This means the outcome of a conflict is predictable and
auditable without inspecting model internals.
Combined with the full audit trail maintained by the
logging middleware, SAI produces a complete, verifiable
record of every authority decision made during a
mission.
 
These three properties together address a gap in
current LLM-agent deployment practice: the absence of
a \emph{verifiable human authority layer} that is
architecturally prior to model inference~\cite{raptis2025agentic}.
SAI provides this layer without requiring model
modification, making it applicable to any instruction-tuned model that supports MCP tool use~\cite{li2025survey}.

\begin{table}[t]
\caption{Scenario details: eight planning failure modes.}
\label{tab:scenarios}
\renewcommand{\arraystretch}{1.1}
\begin{tabularx}{\columnwidth}{@{}clXl@{}}
\toprule
\# & Failure mode & Setup summary & SAI class \\
\midrule
S1 & Priority mis-order & 2 robots; high-priority medical task far from spawn & Priority \\
S2 & Zone distribution & 3 robots clustered left; must spread across both zones & Strategy \\
S3 & Compound & 3 robots; satisfy priority + zone rules simultaneously & P + S \\
S4 & Late interrupt & SearchTask spawns at tick 10; redirect robot & Priority \\
S5 & Unreachable task & High-priority task in sealed room; must not assign & Constraint \\
S6 & Capability mismatch & 6 robots $\times$ 3 types; 8 tasks with requirements & Constraint \\
S7 & Bottleneck & 6 robots; single-cell corridor; limit concurrency & Strategy \\
S8 & Formation & 5 robots; 2 cargo tasks need $\geq$3 robots; sequence & Strategy \\
\bottomrule
\end{tabularx}
\vspace{-0.15in}
\end{table}
\section{Experimental Evaluation}
\label{sec:evaluation}

\begin{table*}[!htbp]
\centering
\caption{Task success rate comparison between baseline and SAI across eight scenarios.}
\label{tab:sai_results}

\scriptsize
\footnotesize
\renewcommand{\arraystretch}{1.05}

\resizebox{\textwidth}{!}{
\begin{tabular}{lcccccccccccccccc}
\toprule

& \multicolumn{2}{c}{S1}
& \multicolumn{2}{c}{S2}
& \multicolumn{2}{c}{S3}
& \multicolumn{2}{c}{S4}
& \multicolumn{2}{c}{S5}
& \multicolumn{2}{c}{S6}
& \multicolumn{2}{c}{S7}
& \multicolumn{2}{c}{S8} \\[-2pt]

\cmidrule(lr){2-3}
\cmidrule(lr){4-5}
\cmidrule(lr){6-7}
\cmidrule(lr){8-9}
\cmidrule(lr){10-11}
\cmidrule(lr){12-13}
\cmidrule(lr){14-15}
\cmidrule(lr){16-17}

LLM
& Base & SAI
& Base & SAI
& Base & SAI
& Base & SAI
& Base & SAI
& Base & SAI
& Base & SAI
& Base & SAI\\

\midrule

ASI1
& 1.00 & 1.00
& 1.00 & 1.00
& 1.00 & 1.00
& \textbf{0.80} & \cellcolor{red!15}0.00
& 0.00 & \cellcolor{green!15}\textbf{0.75}
& \textbf{1.00} & \cellcolor{red!15}0.88
& \textbf{1.00} & \cellcolor{red!15}0.86
& \textbf{1.00} & \cellcolor{red!15}0.00 \\

Claude-Haiku
& 0.75 & \cellcolor{green!15}\textbf{1.00}
& 1.00 & 1.00
& 1.00 & 1.00
& 0.83 & \cellcolor{green!15}\textbf{1.00}
& 0.75 & 0.75
& 1.00 & 1.00
& 1.00 & 1.00
& \textbf{1.00} & \cellcolor{red!15}0.50 \\

DeepSeek-R1
& 1.00 & 1.00
& 1.00 & 1.00
& 1.00 & 1.00
& 0.80 & \cellcolor{green!15}\textbf{1.00}
& 0.00 & \cellcolor{green!15}\textbf{0.75}
& 1.00 & 1.00
& \textbf{1.00} & --
& 1.00 & 1.00 \\

Gemini-2.0-Flash
& 0.50 & 0.50
& \textbf{1.00} & \cellcolor{red!15}0.50
& 1.00 & 1.00
& 0.80 & \cellcolor{green!15}\textbf{1.00}
& 0.75 & 0.75
& \textbf{1.00} & \cellcolor{red!15}0.88
& 0.43 & \cellcolor{green!15}\textbf{0.57}
& \textbf{1.00} & \cellcolor{red!15}0.50 \\

Gemma-4-26B
& 1.00 & 1.00
& 1.00 & 1.00
& 1.00 & 1.00
& 0.80 & \cellcolor{green!15}\textbf{1.00}
& 0.75 & 0.75
& \textbf{1.00} & \cellcolor{red!15}0.88
& 1.00 & 1.00
& 1.00 & 1.00 \\

GPT-4o
& 1.00 & 1.00
& 1.00 & 1.00
& 1.00 & 1.00
& 0.80 & \cellcolor{green!15}\textbf{0.83}
& 0.75 & 0.75
& 0.88 & 0.88
& \textbf{1.00} & \cellcolor{red!15}0.57
& 1.00 & 1.00 \\

GPT-OSS-20B
& 1.00 & 1.00
& 0.00 & \cellcolor{green!15}\textbf{1.00}
& 0.20 & \cellcolor{green!15}\textbf{1.00}
& \textbf{0.80} & \cellcolor{red!15}0.17
& 0.00 & \cellcolor{green!15}\textbf{0.75}
& 0.63 & \cellcolor{green!15}\textbf{1.00}
& \textbf{1.00} & \cellcolor{red!15}0.57
& 0.00 & \cellcolor{green!15}\textbf{1.00} \\

LLaMA-3.3-70B
& 1.00 & 1.00
& 1.00 & 1.00
& 1.00 & 1.00
& 1.00 & 1.00
& 0.00 & \cellcolor{green!15}\textbf{0.50}
& 0.88 & 0.88
& \textbf{0.43} & \cellcolor{red!15}0.00
& 1.00 & 1.00 \\

Qwen3-32B
& 1.00 & 1.00
& 1.00 & 1.00
& 1.00 & 1.00
& 0.80 & \cellcolor{green!15}\textbf{1.00}
& 0.75 & 0.75
& 0.88 & 0.88
& 0.43 & \cellcolor{green!15}\textbf{0.71}
& 0.00 & \cellcolor{green!15}\textbf{1.00} \\

\bottomrule
\end{tabular}
}
\vspace{2pt}

\footnotesize
\textit{Note:} Green cells indicate improvement and red cells indicate degradation.
Bold values denote the better result between conditions. A dash (–) denotes an excluded run: DeepSeek-R1 under SAI in S7 exhausted its 4,096-token generation budget before producing any assignment.
\end{table*}

\begin{figure*}[b]
\centering
\vspace{-0.1in}
\includegraphics[scale=0.36]{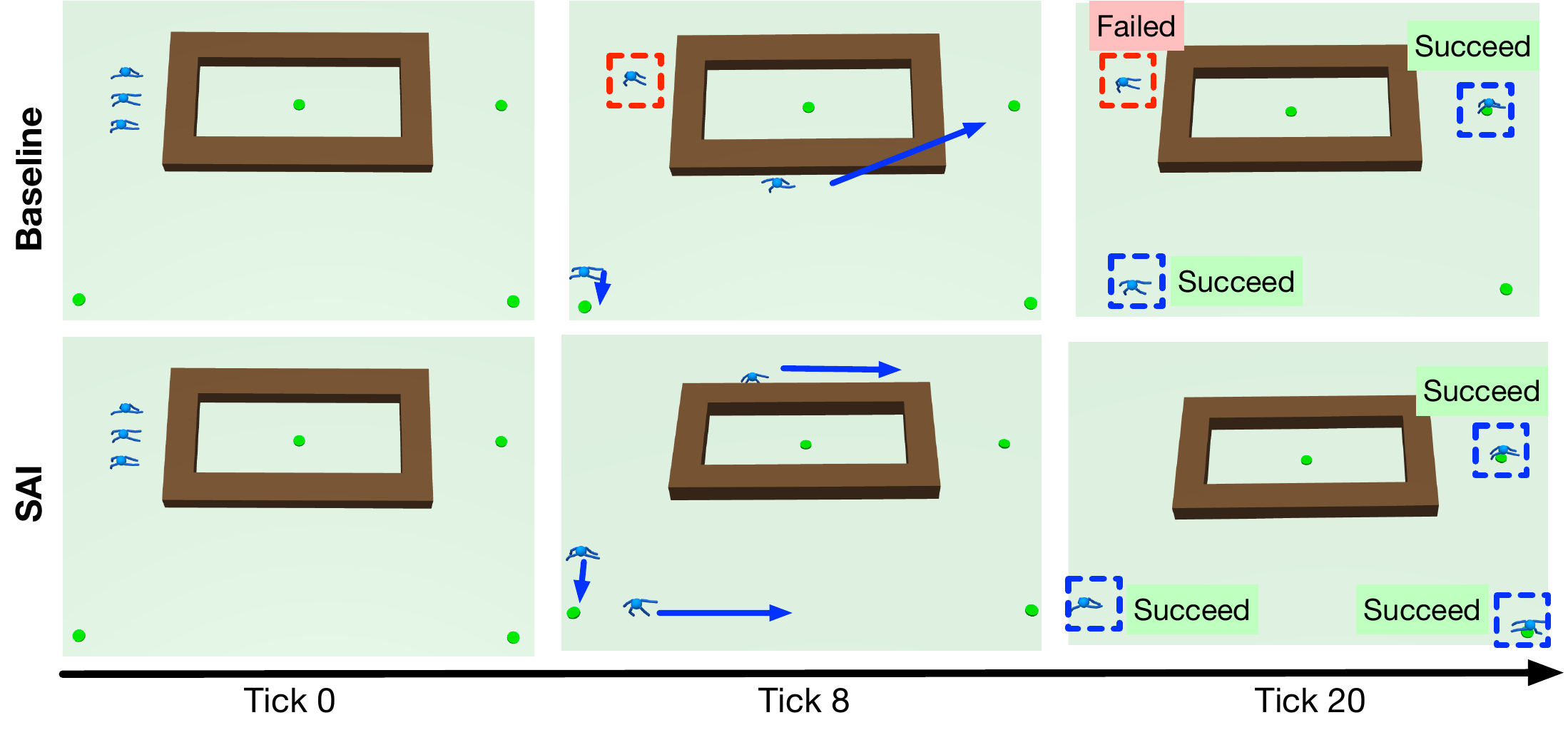}
\caption{Baseline failure vs. SAI under an obstructed task. Robots are colored in blue, and targets are colored in green.}
\label{fig:case}
\end{figure*}

\subsection{Scenario and Setup}
\label{sec:setup}
 
We design eight scenarios each targeting one distinct failure mode (Table~\ref{tab:scenarios}). The Baseline condition gives the LLM no pre-prompting beyond the world state. The SAI condition injects relevant context as a typed SAI override before the first planning epoch. This isolates each failure mode so that performance differences are attributable to the override mechanism, not task difficulty.
 
Nine LLMs are evaluated: GPT-4o~\cite{gpt4o2024}, DeepSeek-R1~\cite{guo2025deepseek}, Qwen3-32B~\cite{yang2025qwen3}, Claude-Haiku~\cite{claudehaiku2024}, Gemini-2.0-Flash~\cite{gemini2024}, LLaMA-3.3-70B~\cite{grattafiori2024llama}, Gemma-4-26B~\cite{team2024gemma}, GPT-OSS-20B~\cite{gptoss2025}, and ASI1~\cite{asi12025}. Each model $\times$ condition pair is run five times on 100-tick simulations. All eight scenarios are instantiated in a MuJoCo-based grid-world simulator, where each robot is modeled as a kinematic agent with discrete move and interact actions, and task completion is detected automatically via proximity and 
state-change triggers.

\subsection{Override Correctness}
\label{sec:rates}

Table~\ref{tab:sai_results} reports task success rates across all models and scenarios. SAI recovers completions for models with \emph{partial} failures, for instance, Claude-Haiku improves from 0.75 to 1.00 in S1, and GPT-OSS-20B recovers from 0.00 to 1.00 in both S2 and S3, but cannot repair systematic omission: Gemini-2.0-Flash remains at 0.50 in S1 under both conditions. Notably, Gemini-2.0-Flash also regresses from 1.00 to 0.50 in S2 under SAI, confirming that structured injection can disrupt models already performing well.

In S1--S3, most models operate near ceiling under baseline, leaving little room for SAI to improve. S4 (late interrupt) exposes the sharpest divergence: SAI improves DeepSeek-R1, Gemini, Gemma-4, and Claude-Haiku (each $+$0.17--$+$0.20), while severely degrading ASI1 (0.80$\to$0.00) and GPT-OSS (0.80$\to$0.17), indicating that sensitivity to structured context injection is model-dependent rather than task-difficulty-dependent.

The constraint-heavy scenarios (S5--S8) amplify these heterogeneous effects. SAI yields large gains in S5 for ASI1, GPT-OSS, and DeepSeek-R1 (all 0.00$\to$0.75), but leaves models already at 0.75 baseline unchanged. In S8, GPT-OSS and Qwen3 both recover from 0.00 to 1.00, while ASI1 collapses in the same scenario (1.00$\to$0.00), and GPT-4o and LLaMA drop sharply in S7 ($-$0.43 each). DeepSeek-R1's S7 result under SAI is excluded: its single planning call exhausted the 4,096-token output limit before producing any assignments, as the combined authority context and seven-task state inflated the reasoning chain beyond the model's generation budget. Overall, SAI acts as a high-variance intervention, which is beneficial in under-constrained or previously failed settings, but destabilising in tightly coupled multi-constraint regimes.

\textbf{\textit{Case Study: Persistent Task Misallocation.}} Figure~\ref{fig:case} shows a case study for 3 ant robots in scenario 5. At tick~0, both Baseline and SAI begin from identical robot configurations. By tick~8, the Baseline planner has already collapsed into a persistent invalid allocation: the upper-left robot remains assigned to an obstructed and unreachable task, while the remaining robots continue normal navigation.
In contrast, SAI redirects all robots away from the blocked objective through structured authority constraints, producing a globally feasible allocation.
By tick~20, SAI successfully completes the scenario, whereas the Baseline system remains deadlocked with the same robot still targeting the impossible task. This case illustrates how persistent authority context prevents long-horizon assignment failures that the Baseline planner cannot recover from.

\subsection{Authority Persistence}
\label{sec:results-timeshare}

\begin{figure}[t]
\centering
\includegraphics[width=1.1\columnwidth]{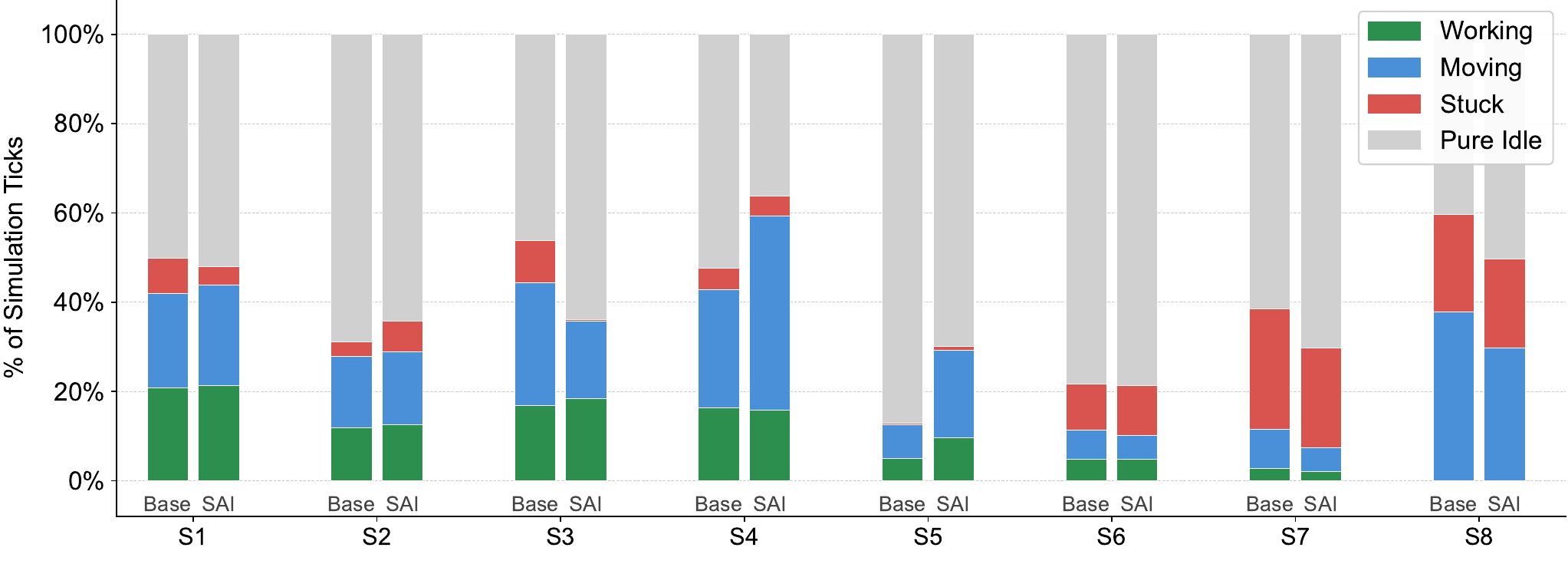}
\caption{Robot time-state distribution across scenarios.}
\label{fig:robot_timeshare}
\end{figure}

\begin{figure}[t]
\centering
\includegraphics[width=1.1\columnwidth]{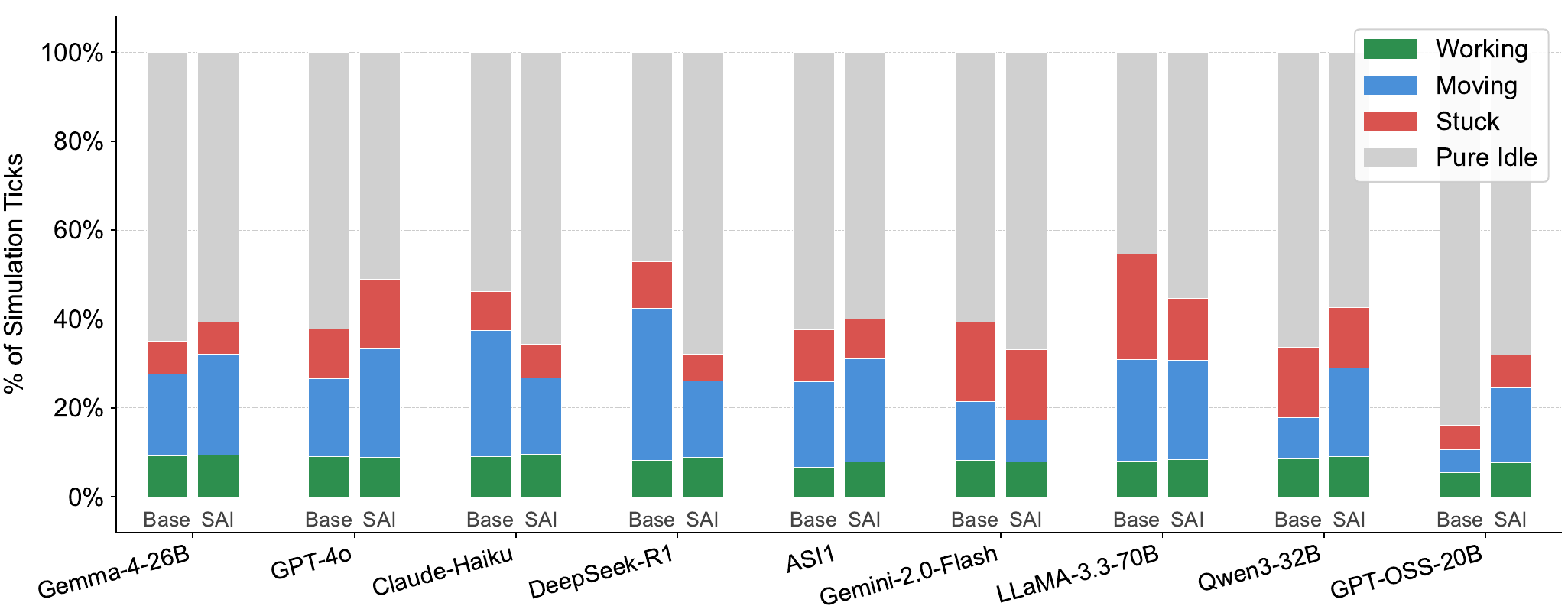}
\caption{Robot time-state distribution across LLMs.}
\label{fig:robot_llm_timeshare}
\end{figure}

Figure~\ref{fig:robot_timeshare} breaks down robot time across
four mutually exclusive states averaged over all nine LLMs per scenario-condition pair, including Working (executing a task on-site), Moving (traveling toward an assignment), Stuck (assigned but physically blocked for that tick by a collision or obstacle), and Pure Idle (no current assignment).
Pure Idle dominates in every scenario ($36$--$87\%$), a structural consequence of the batched planning loop and post-completion dead ticks; high idle therefore coexists with perfect success rates and does not imply planning failure.
The informative cross-condition signals are in Moving and Stuck.
SAI sharply increases Moving in S4 ($26\%\!\to\!44\%$) as the late interrupt forces robots to re-navigate to reprioritised targets, and nearly eliminates Stuck in S3 ($9\%\!\to\!0\%$) where structured zone constraints resolve baseline spatial conflicts.
S7 carries the highest Stuck burden overall ($\approx\!27\%$ baseline), while S5 shows the broadest activation increase ($13\%\!\to\!30\%$), consistent with the completion gains in Table~\ref{tab:sai_results}.

Figure~\ref{fig:robot_llm_timeshare} applies the same decomposition by LLM across all eight scenarios.
Working time is near-uniform ($6$--$10\%$) across all models, confirming that on-task execution is model-agnostic; planning quality differences surface in Moving and Stuck instead.
GPT-OSS-20B records the lowest baseline active time ($\approx\!16\%$), which SAI doubles to $\approx\!32\%$, consistent with its collapsed success rates in S2, S3, and S5.
DeepSeek-R1 shows the sharpest SAI reduction in Moving ($34\%\!\to\!17\%$), indicating more spatially precise assignments under structured authority. LLaMA-3.3-70B's Stuck fraction falls from $24\%$ to $14\%$ under SAI, while GPT-4o is a notable counter-example whose Stuck fraction \emph{increases} ($11\%\!\to\!16\%$), showing that structured directives can introduce routing conflicts even when task completion improves.

\subsection{Token efficiency}
\label{sec:results-tokens}

\begin{figure}[t]
\centering
\includegraphics[width=0.85\columnwidth]{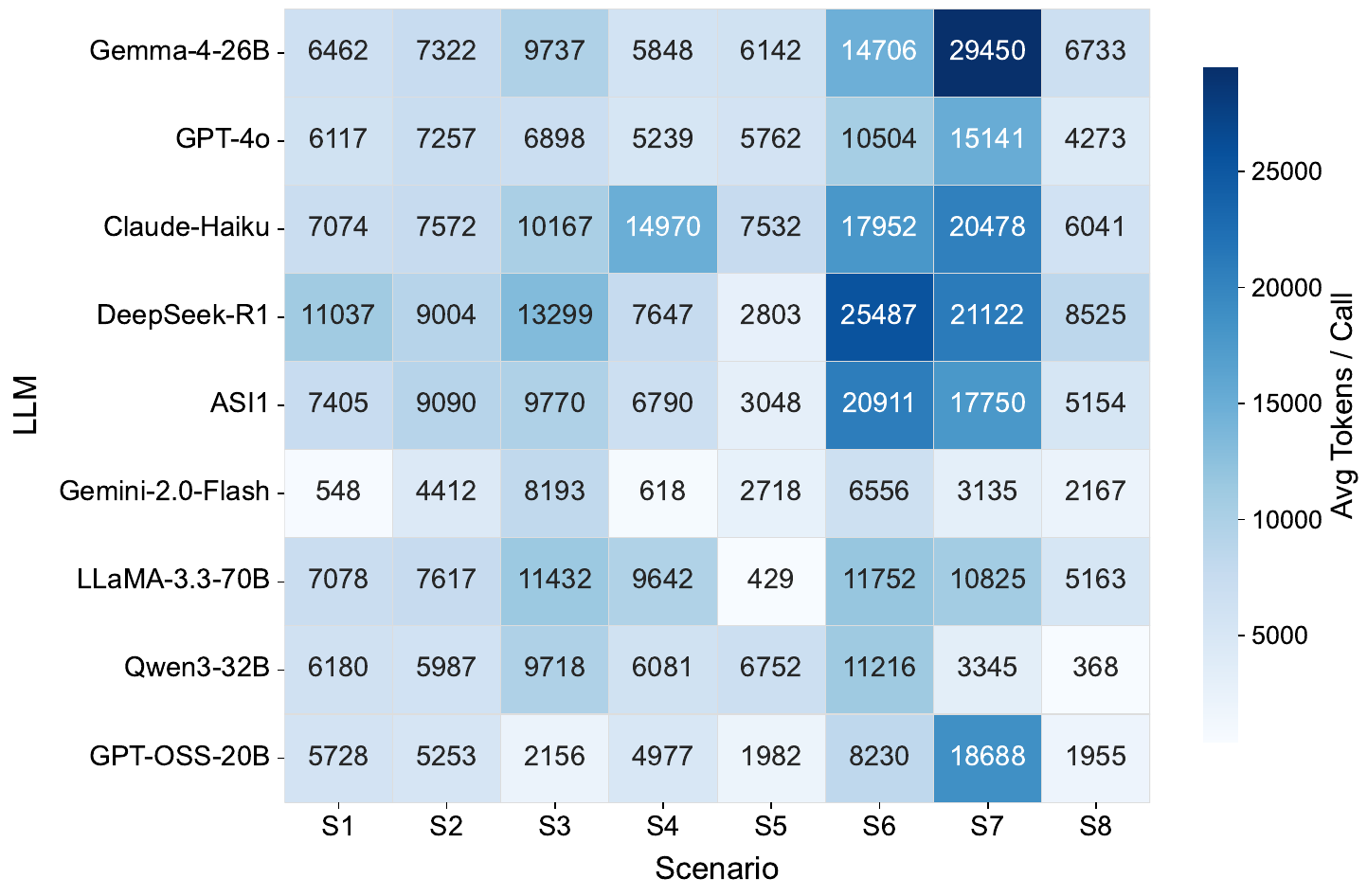}
\caption{The token usage of Baseline.}
\label{fig:tokens_baseline}
\end{figure}

\begin{figure}[t]
\centering
\includegraphics[width=0.85\columnwidth]{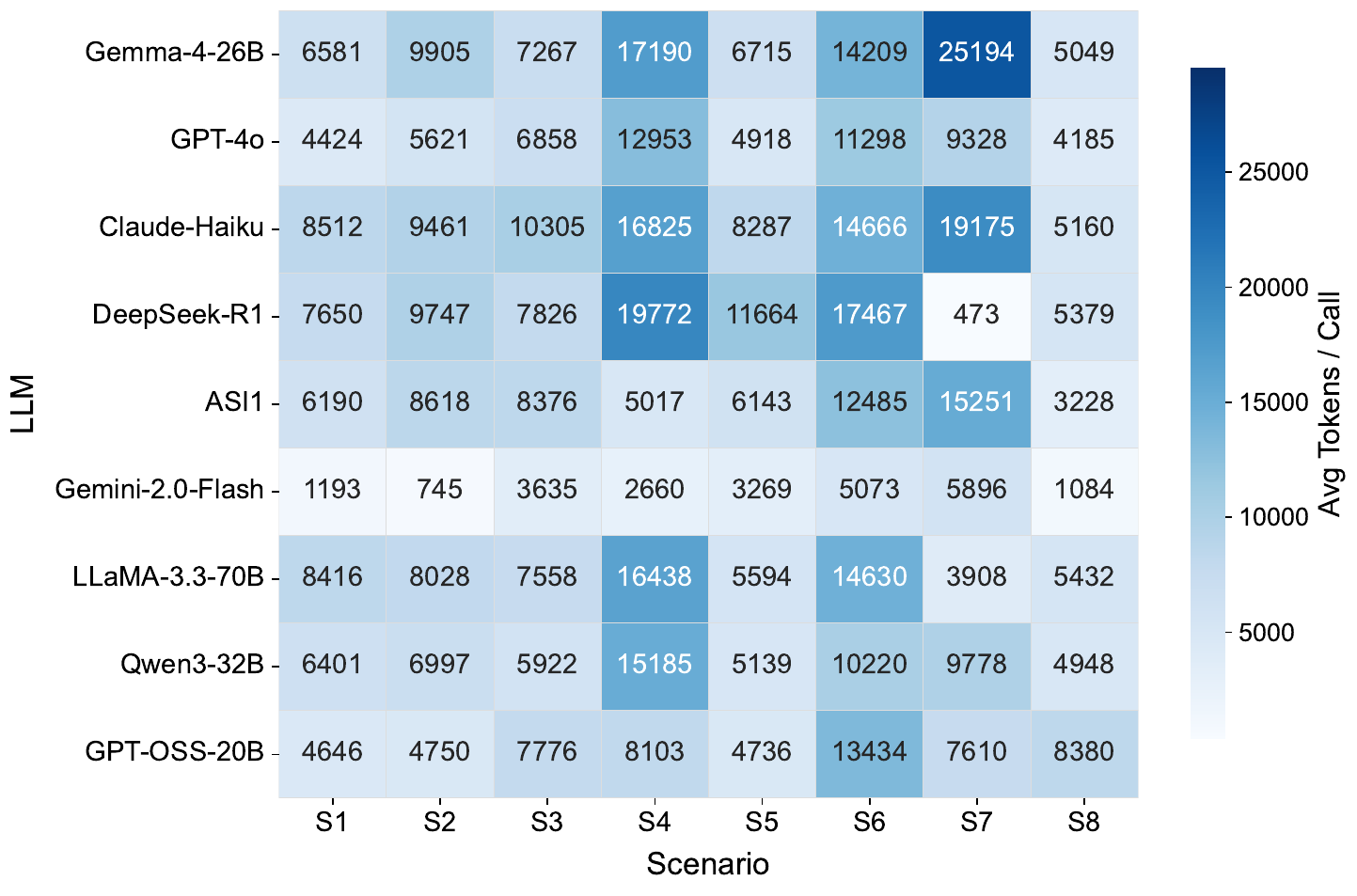}
\caption{The token usage of SAI. (The DeepSeek-R1 / S7 cell reflects a generation-budget timeout, not a token saving; see Section IV-D.)}
\label{fig:tokens_sai}
\end{figure}

Figure~\ref{fig:tokens_baseline} and Figure~\ref{fig:tokens_sai} report average input tokens per agent call across scenarios. First, the two strongest models consume fewer tokens under SAI. GPT-4o and DeepSeek-R1 show consistent reductions across most scenarios. The most pronounced reductions occur in S6–S7: DeepSeek-R1 falls from 25,487 to 17,467 in S6 ($-$31\%), and GPT-4o from 15,141 to 9,328 in S7 ($-$38\%). DeepSeek-R1's S7 value (473) is excluded as a generation-budget timeout, not a saving.
We attribute these reductions to the authority context providing sufficiently precise constraints that the model reaches valid assignments in fewer iterations, reducing the accumulated tool-call history that dominates prompt length in later epochs.

Second, the remaining models show mixed or increased consumption. Claude-Haiku, LLaMA-3.3-70B, and Qwen3-32B exhibit token increases under SAI in several scenarios (Qwen3-32B rises from 3{,}345 to 9{,}778 in S7) suggesting these models do not consistently exploit structured directives to prune their planning search, paying a token cost without a corresponding planning benefit.
Gemini-2.0-Flash records the lowest absolute counts in both conditions, but its task-completion failures indicate that low consumption here reflects insufficient planning depth rather than efficiency.
S4, S6, and S7 are the high-cost scenarios in both heatmaps regardless of model or condition, confirming that scenario-driven complexity (i.e., late interrupts, capability matching, bottleneck congestion) sets a token floor that SAI does not eliminate. Token efficiency and task-completion benefit are correlated: GPT-4o and DeepSeek-R1, which gain most from SAI in completion rate, also gain most in token economy.

\section{Findings and Discussion}
\label{sec:discussion}

\noindent\textbf{Finding 1: SAI benefit is stratified by observed capability.}
\textit{Some models gain across every dimension; others incur token overhead without commensurate planning gains, and model scale does not predict which is which.}

The models fall into three empirical tiers.
Tier~1 (GPT-4o, DeepSeek-R1) exploits the authority context to prune its
planning search, achieving token reductions of up to 38\% and sharper spatial
assignment precision.
Tier~2 (GPT-OSS-20B) recovers large completion gains in previously failing
scenarios but remains sensitive to context complexity under late interrupts.
Tier~3 (LLaMA-3.3-70B, Claude-Haiku, Qwen3-32B) shows mixed completion effects
while consistently incurring token overhead.
Tier membership tracks neither parameter count nor model family: the 20B GPT-OSS and the 70B LLaMA-3.3 land in different tiers, and Qwen3-32B recovers
from 0.00 to 1.00 in S8 yet rises from 3,345 to 9,778 tokens in S7 without a
corresponding benefit.
A further boundary case is DeepSeek-R1 in S7, where the authority context
inflated the reasoning chain beyond its 4,096-token budget, producing complete
planning failure---a risk not captured by standard tool-use benchmarks.

For operators, neither parameter count nor model family can substitute for
direct measurement: generation budget must be verified per model against the
expected authority-context length, making a lightweight qualification benchmark
a necessary pre-deployment gate.

\smallskip
\noindent\textbf{Finding 2: MCP as an authority layer, not just a tool bus.}
\textit{SAI's asymmetric callability transforms MCP from a bidirectional tool bus into a one-directional authority channel: override tools are accessible only to the supervision interface, never to the LLM.} Because schema validation is prior to LLM inference, malformed or adversarially injected directives are rejected before influencing planning---a guarantee that prompt-convention-based HitL designs~\cite{li2025hmcf,liu2023sirius} cannot provide at the protocol level.
 
This architectural inversion has direct implications for agentic system security. Prompt injection attacks, where malicious content in tool outputs attempts to hijack agent behavior, are an active threat in deployed LLM systems.
In LLM-driven network management or cloud orchestration, a successful injection that removes a Constraint override could silently disable a traffic engineering rule or bypass a resource quota.
SAI's server-side validation firewall eliminates this vector: the authority context is writable only via a credentialed supervision interface, making tamper resistance an architectural property rather than a prompt-engineering convention.
Furthermore, because the precedence rule ($C\succ P\succ S$) is evaluated deterministically at prompt-construction time, every authority decision is fully auditable without inspecting model internals. This is a property essential for regulated deployment contexts such as network security operations and industrial automation.
 
\smallskip
\noindent\textbf{Finding 3: Stuck-tick rate as a planning quality diagnostic.}
Working time remains near-uniform across all models (6--10\%); planning
quality differences manifest in the stuck fraction. LLaMA-3.3-70B's stuck fraction falls from 24\% to 14\% under SAI, whereas GPT-4o's rises from 11\% to 16\% despite unchanged or improved completion in most scenarios, closely mirroring the divergence in their assignment quality. 
Task completion rate is a coarse metric that can mask persistent planning pathologies: a planner repeatedly assigning agents to infeasible targets will eventually complete all tasks while wasting substantial time in blocked and transit states.
Stuck-tick rate, the fraction of ticks in which any agent is assigned but physically blocked, that directly quantifies this waste.
We propose it as a standard diagnostic complement to completion-based measures in LLM-driven multi-agent evaluation, noting its low instrumentation cost: it requires only a per-agent state flag per tick, available in any environment with collision or contention detection.
 
\smallskip
\noindent\textbf{Limitations and Future Work.}
The current design assumes a single trusted supervisor; multi-operator deployments would require an arbitration layer to resolve conflicting overrides before the class-level precedence rule is applied. Domain-specific semantics may necessitate taxonomy extensions beyond the three current classes. Beyond these structural limitations, SAI introduces a non-trivial prompt overhead that can destabilize models already operating at or near ceiling performance, as evidenced by Gemini-2.0-Flash's regression in S2 and the drops observed for GPT-4o and LLaMA-3.3-70B in S7. These degradations suggest that authority context injection can introduce competing directive signals that interfere with well-calibrated planners, and that the benefit–cost balance of structured injection is sensitive to baseline model priors and scenario constraint density; mitigating this may require adaptive context compression or selective directive rendering conditioned on model capacity. Furthermore, transferring SAI to physical deployments introduces communication latency, sensor noise, and partial observability not modeled here, and physical-platform validation remains necessary before safety-critical adoption. Future work will also explore SAI as a runtime governance layer for LLM-orchestrated network and cloud systems, where the authority channel pattern is directly applicable to operator-driven policy enforcement over autonomous infrastructure agents.

\section{Conclusion}
\label{sec:conclusion}

We presented Structured Authority Injection (SAI), a protocol that routes human operator corrections through typed MCP tool calls and maintains a persistent, ranked authority context prepended to every LLM prompt. Evaluated across nine LLMs and eight systematically designed scenarios, SAI eliminates the ambiguity and authority-erosion failure modes that plague free-text override approaches, while revealing that the authority context itself imposes a generation-budget requirement that must be verified before deployment. By repurposing MCP's schema-validation machinery as a one-directional human authority layer as tamper-resistant and architecturally prior to LLM inference, SAI offers a transferable design pattern for any human-robot interaction context where operator corrections must be unambiguous, immediate, and persistent.


%
%
\bibliographystyle{IEEEtran}
\bibliography{reference}

\end{document}